\documentclass[letterpaper, 10 pt, conference]{ieeeconf}  

\IEEEoverridecommandlockouts                              

\usepackage{graphics} 
\usepackage{epsfig} 
\usepackage{mathptmx} 
\usepackage{times} 
\usepackage{amsmath} 
\usepackage{amssymb}  
\usepackage{gensymb}
\usepackage{xcolor}
\usepackage{mathrsfs}
\usepackage{amstext}
\usepackage{multirow}
\makeatletter
\let\NAT@parse\undefined
\makeatother
\def\secref#1{Sec.~\ref{#1}}
\def\figref#1{Fig.~\ref{#1}}
\def\tabref#1{Tab.~\ref{#1}}
\def\eqref#1{Eq.~(\ref{#1})}

\usepackage{xcolor}

\DeclareMathAlphabet\mathbfcal{OMS}{cmsy}{b}{n}
\usepackage{amsmath}
\usepackage{calc}
\usepackage{bm}
\usepackage{xspace}
\usepackage{tabularx}
\usepackage[numbers]{natbib}
\newcolumntype{L}[1]{>{\raggedright\arraybackslash}p{#1}}
\newcolumntype{C}[1]{>{\centering\arraybackslash}p{#1}}
\newcolumntype{R}[1]{>{\raggedleft\arraybackslash}p{#1}}
\usepackage{subfigure}
\usepackage{textcomp}
\title{\LARGE \bf
Deep Semantic Classification for 3D LiDAR Data
}

\author{Ayush Dewan \and Gabriel L. Oliveira \and Wolfram Burgard%
\thanks{All authors are with the Department of Computer Science at the University of Freiburg, Germany. This work has been supported by the European Commission under the grant numbers FP7-610532-SQUIRREL}%
}

\begin{document}

\thispagestyle{empty}
\pagestyle{empty}

\onecolumn
{\Large

\textcopyright IEEE. Personal use of this material is permitted. Permission from IEEE must be obtained
for all other uses, in any current or future media, including reprinting/republishing this material
for advertising or promotional purposes, creating new collective works, for resale or redistribution
to servers or lists, or reuse of any copyrighted component of this work in other works.\\

%
Pre-print of article that will appear at the 2017 IEEE International Conference on Intelligent
Robots and Systems.\\

Please cite this paper as:\\
A. Dewan, G. Oliveira and W. Burgard, "Deep Semantic Classification for 3D LiDAR Data" in Intelligent Robots and Systems (IROS), 2017 IEEE/RSJ International Conference on. IEEE, 2017.\\

bibtex:\\
@inproceedings$\lbrace$ dewan17iros,\\ 
author = $\lbrace$ Ayush Dewan, Gabriel L. Oliveira and Wolfram Burgard$\rbrace$,\\ 
booktitle = $\lbrace$ Intelligent Robots and Systems (IROS), 2017 IEEE/RSJ International Conference on$\rbrace$,\\ 
organization = $\lbrace$IEEE$\rbrace$,\\ 
title = $\lbrace$ $\lbrace$Deep Semantic Classification for 3D LiDAR Data$\rbrace$ $\rbrace$,\\ 
year = $\lbrace$2017$\rbrace$ \\
$\rbrace$
}

\twocolumn

\maketitle
\begin{abstract}
Robots are expected to operate autonomously in dynamic environments. Understanding the underlying dynamic characteristics of objects is a key enabler for achieving this goal. In this paper, we propose a method for pointwise semantic classification of 3D LiDAR data into three classes: non-movable, movable and dynamic. 
We concentrate on understanding these specific semantics because they characterize important information required for an autonomous system. Non-movable points in the scene belong to unchanging segments of the environment, whereas the remaining classes corresponds to the changing parts of the scene. The difference between the movable and dynamic class is their motion state. The dynamic points can be perceived as moving, whereas movable objects can move, but are perceived as static. To learn the distinction between movable and non-movable points in the environment, we introduce an approach based on deep neural network and for detecting the dynamic points, we estimate pointwise motion. We propose a Bayes filter framework for combining the learned semantic cues with the motion cues to infer the required semantic classification. In extensive experiments, we compare our approach with other methods on a standard benchmark dataset and report competitive results in comparison to the existing state-of-the-art. Furthermore, we show an improvement in the classification of points by combining the semantic cues retrieved from the neural network with the motion cues.
\end{abstract}

\section{INTRODUCTION}
One of the vital goals in mobile robotics is to develop a system that is aware of the dynamics of the environment. If the environment changes over time, the system should be capable of handling these changes. 
In this paper, we present an approach for pointwise semantic classification of a 3D LiDAR scan into three classes: \textit{non-movable, movable and dynamic}. Segments in the environment having non-zero motion are considered dynamic, a region which is expected to remain unchanged for long periods of time is considered non-movable, whereas the frequently changing segments of the environment is considered movable. Each of these classes entail important information. Classifying the points as dynamic facilitates robust path planning and obstacle avoidance, whereas the information about the non-movable and movable points can allow uninterrupted navigation for long periods of time.

\begin{figure}[t]
  \centering
    \fbox{\includegraphics[width=0.40\textwidth]{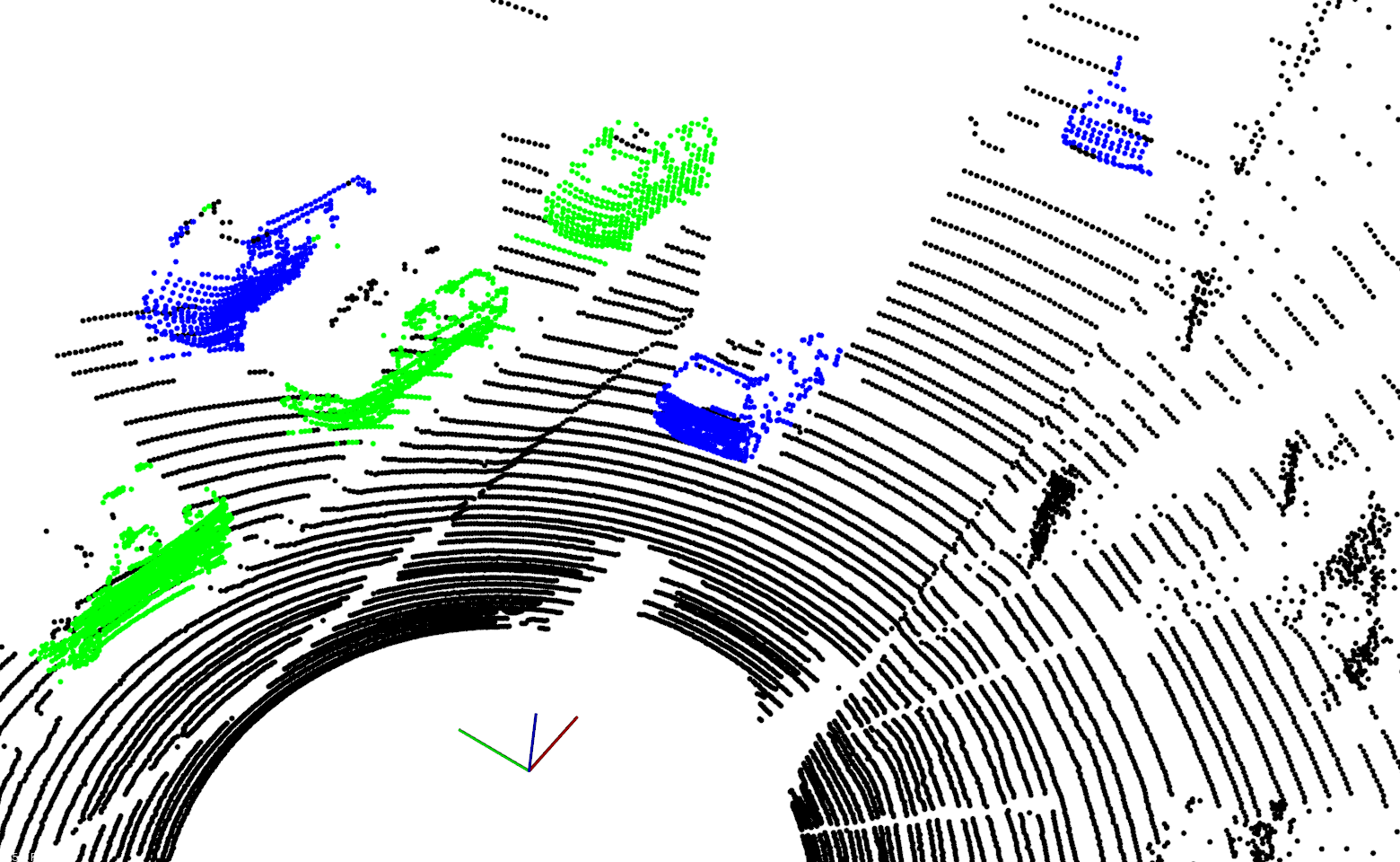}}
    \caption{Semantic classification of a 3D LiDAR scan. \textit{Non-movable} points belonging to building, vegetation and road are colored black. Points on parked vehicles, correctly classified as \textit{movable} are shown in green, whereas the points on moving vehicles are classified as \textit{dynamic} and are shown in blue.}
    \label{fig:summary}
\end{figure}

To achieve the desired objective, we use a Convolutional Neural Network (CNN)~\cite{VGG,He2015,huang2016densely} for understanding the distinction between
movable and non-movable points. For our approach, we employ a particular type of CNNs called up-convolutional networks~\cite{OB16b}. They are fully convolutional architectures capable of producing dense predictions for a high-resolution input. The input to our network is a set of three channel 2D images generated by unwrapping $360\degree$ 3D LiDAR data onto a spherical 2D plane and the output is the \textit{objectness} score, where a high score corresponds to the movable class. Similarly, we estimate the \textit{dynamicity} score for a point by first calculating pointwise 6D motion using our previous method~\cite{dewan2016rigid} and then comparing the estimated motion with the odometry to calculate the score. We combine the two scores in a Bayes filter framework for improving the classification especially for dynamic points. Furthermore, our filter incorporates previous measurements, which makes the classification more robust. In~\figref{fig:summary} we show the classification results of our method. Black points represent non-movable points, whereas movable and dynamic points are shown in green and blue color respectively. 

\begin{figure*}[!t]
  \centering
    \fbox{\includegraphics[width=0.95\textwidth]{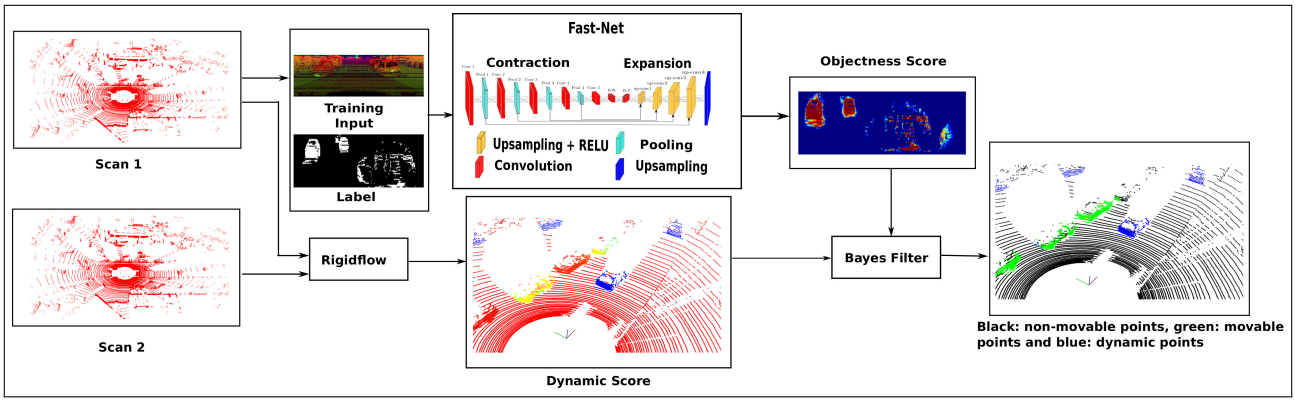}}
    \caption{An overview of the proposed method. The input is two LiDAR scans, where the first scan is converted to a 2D image and is processed by the neural network for estimating the \textit{objectness} score. Our RigidFlow approach uses both scans and calculates the \textit{dynamicity} score. The \textit{objectness} and the \textit{dynamicity} score is then processed by the Bayes filter approach for estimating the desired pointwise semantic classification of the scene.}
    \label{fig:overview}
\end{figure*}

Other methods~\cite{reddy2014semantic,fan2016semantic} for similar semantic classification have been proposed for RGB images, however, a method solely relying on range data does not exist according to our best knowledge. For LiDAR data, separate methods exists for both object detection~\cite{chen2016multi,li2016vehicle,engelcke2016vote3deep,XiaozhiCVPR16} and for distinguishing between static and dynamic objects in the scene~\cite{dewan16icra,moosmann2013joint,pomerleau2014long}. The two main differences between our method and the other object detection methods is that the output of our method is a pointwise \textit{objectness} score, whereas other methods concentrate on calculating object proposals and predict a bounding box for the object. Since our objective is pointwise classification, the need for estimating a bounding box is alleviated as a pointwise score currently suffices. The second difference is that we utilize the complete 360$\degree$ field of view (FOV) of LiDAR for training our network in contrast to other methods which only use the points that overlap with the FOV of the front camera. 

The main contribution of our work is a method for semantic classification of a LiDAR scan for learning the distinction between non-movable, movable and dynamic parts of the scene. As mentioned above, these three classes encapsulate information which is critical for robust autonomous robotic system. A method for learning the same classes in LiDAR scans has not been proposed before, even though different methods exists for learning other semantic level information. Unlike other existing methods, we use the complete range of the LiDAR data. For training the neural network we use the KITTI object benchmark~\cite{Geiger2012CVPR} and compare our results on this benchmark with the other methods. We also
test our approach on the dataset by~\citeauthor{moosmann2013joint}~\cite{moosmann2013joint}.

\section{Related Works}
We first discuss methods which have been proposed for similar classification objectives. Then, we discuss other methods proposed especially for classification in LiDAR scans and briefly discuss the RGB image based methods.

For semantic motion segmentation in images, \citeauthor{reddy2014semantic}~\cite{reddy2014semantic} proposed a dense CRF based method, where they combine semantic, geometric and motion constraints for joint pixel-wise semantic and motion labeling. Similar to them~\citet{fan2016semantic} proposed a neural network based method. Their method is closest to our approach as they also combine motion cues with the object information. For retrieving the object level semantics, they use a deep neural network~\cite{kendall2015}. Both of these methods show results on the KITTI sceneflow benchmark for which ground truth is only provided for the images, thus making a direct comparison difficult. However, we compare the
performance of our neural network with the network used by~\citet{fan2016semantic}.

For LiDAR data, a method with similar classification objectives does not exist, however different methods for semantic segmentation~\cite{zelener2016cnn,dohan2015learning,wang2012could}, object detection~\cite{chen2016multi,li2016vehicle,engelcke2016vote3deep} and moving object segmentation~\cite{dewan16icra,moosmann2013joint,pomerleau2014long} have been proposed. Targeting semantic segmentation in 3D LiDAR data, Wang~\emph{et al.} proposed a method~\cite{wang2012could} for segmenting movable objects. More recently \citeauthor{zelener2016cnn} proposed a method~\cite{zelener2016cnn} for object segmentation, primarily concentrating on objects with missing points and \citeauthor{dohan2015learning}~\cite{dohan2015learning} discusses a method for hierarchical semantic segmentation of a LiDAR scan. These methods report results on different datasets and since we use the KITTI object benchmark for our approach we restrict our comparison to other recent methods that use the same benchmark.
 
For object detection,  \citeauthor{engelcke2016vote3deep}~\cite{engelcke2016vote3deep} extends their previous work~\cite{wang2015voting} and propose a CNN based method for detecting objects in 3D LiDAR data. \citeauthor{li2016vehicle} proposed a Fully Convolutional Network based method~\cite{li2016vehicle} for detecting objects, where they use two channel (depth + height) 2D images for training the network and estimate 3D object proposals. 
The most recent approach for detecting objects in LiDAR scans is proposed by \citeauthor{chen2016multi}~\cite{chen2016multi}. Their method leverages over both multiple view point information (front camera view + bird eye view) and multiple modalities (LiDAR + RGB). They use a region based proposal network for fusing different sources of information and also estimate 3D object proposals. For RGB images, approaches by \citeauthor{XiaozhiCVPR16}~\cite{XiaozhiCVPR16,chen20153d} are the two recent methods for object detection. In All of these methods, the neural network is trained for estimating bounding boxes for object detection, whereas, our network is trained for estimating pointwise \textit{objectness} score; the information necessary for pointwise classification. In the results section, we discuss these differences in detail and present comparative results.

Methods proposed for dynamic object detection include our previous work~\cite{dewan16icra} and other methods~\cite{moosmann2013joint,pomerleau2014long,wang2015model}. Our previous method and~\cite{wang2015model} are model free methods for detection and tracking in 3D and 2D LiDAR scans respectively. For detecting dynamic points in a scene,  \citeauthor{pomerleau2014long}~\cite{pomerleau2014long} proposed a method that relies on a visibility assumption, i.e., the scene behind the object is observed, if an object moves. To leverage over this information, they compare an incoming scan with a global map and detect dynamic points. For tracking and mapping of moving objects a method was proposed by \citeauthor{moosmann2013joint}~\cite{moosmann2013joint}. The main difference between these methods and our approach is that we perform pointwise classification and these methods reason at object level.

\section{Approach}
In this paper, we propose a method for pointwise semantic classification of a 3D LiDAR scan. The points are classified into three classes: non-movable, movable and dynamic. In~\figref{fig:overview} we illustrate a detailed overview of our approach. The input to our approach consists of two consecutive 3D LiDAR scans. The first scan is converted into a three-channel 2D image, where the first channel holds the range values and the second and third channel holds the intensity and height values respectively. The image is  processed by an up-convolutional network called Fast-Net~\cite{OB16b}. The output of the network is the pointwise \textit{objectness} score. Since, in our approach points on an object are considered movable, the term object is used synonymously for movable class. For calculating the \textit{dynamicity} score, our approach requires two consecutive scans. As a first step, we estimate pointwise motion using our RigidFlow~\cite{dewan2016rigid} approach. The estimated motion is then compared with the odometry to calculate the \textit{dynamicity} score. These scores are provided to the Bayes filter framework for estimating the pointwise semantic classification.

\subsection{Object Classification}

Up-convolutional networks are becoming the foremost choice of architectures for 
semantic segmentation tasks based on their recent 
success~\cite{kendall2015,long_shelhamer_fcn,OB16b}. These methods are capable 
of processing images of arbitrary size, are computationally efficient and 
provide the capability of end-to-end training. Up-convolutional networks have 
two main parts: contractive and expansive. The contractive part is a 
classification architecture, for example AlexNet~\cite{alexnet} or 
VGG~\cite{VGG}. They are capable of producing a dense prediction for a 
high-resolution input. However, for a low-resolution output of the contractive 
part, the segmentation mask is not capable of providing the descriptiveness 
necessary for majority of semantic segmentation tasks.  The expansive part 
subdues this limitation by producing an input size output through the 
multi-stage refinement process. Each refinement stage consists of an upsampling 
and a convolution operation of a low-resolution input, followed by the fusion of 
the up-convolved filters with the previous pooling layers. The motivation of 
this operation is to increase the finer details of the segmentation mask at each 
refinement stage by including the local information from pooling.
 
In our approach we use the architecture called Fast-Net~\cite{OB16b} (see~\figref{fig:overview}). It is an up-convolutional network designed for providing near real-time performance. More technical details and a detailed explanation of the 
architecture is described in~\cite{OB16b}. 
\subsection{Training Input}
For training our network we use the KITTI object benchmark. The network is trained for classifying points on \textit{cars} as movable. The input to our network are three channel 2D images and the corresponding ground truth labels. The 2D images are generated by projecting the 3D data onto a 2D point map. The resolution of the image is $64\times870$. Each channel in an image represents a different modality. First channel holds the range values, second channel holds the intensity values, corresponding to the surface reflectance and the third channel holds the height values for providing geometric information. The KITTI benchmark provides ground truth bounding boxes for the objects in front of the camera, even though the LiDAR scanner has 360$\degree$ FOV. To utilize the complete LiDAR information we use our tracking approach~\cite{dewan16icra} for labeling the objects that are behind the camera by propagating the bounding boxes from front of the camera.           
\subsubsection{Training} 

Our approach is modeled as a binary segmentation problem and the goal is to predict the \textit{objectness} score required for distinguishing between movable and non-movable points. We define a set of training images $\mathbfcal{T} =$ ${(\mathit{X}_{n}, \mathit{Y}_{n}), n=1,\dots, N}$, where $\mathit{X_n} = \lbrace x_k,k=1,\dots,|\mathit{X_n}|\rbrace$ is a set of pixels in an example input image and $\mathit{Y}_{n} = \lbrace y_k,k=1,\dots, |\mathit{Y_n}|\rbrace$ is the corresponding ground truth, where $y_k=\lbrace0,1\rbrace$.  The activation function of our model is defined as $f(x_{k},\theta)$, where $\theta$ is our network model parameters. The network learns the features by minimizing the cross-entropy(\textit{softmax}) loss in~\eqref{eq:entopyloss} and the final weights $\theta^*$ are estimated by minimizing the loss over all the pixels as shown in~\eqref{eq:deeploss}.  

\begin{equation} 
\mathbfcal{L}(p,q)=- \sum\limits_{c\in\{0,1\}} p_{c} \log{q_{c}} \label{eq:entopyloss} 
\end{equation} 
\begin{equation} 
\theta^{*} = \underset{\theta}{\operatorname{argmin}} \sum_{k=1}^{N\times|\mathit{X_n}|} \mathbfcal{L}\left ( f\left ( x_k,\theta \right ),y_k \right )\label{eq:deeploss} 
\end{equation}

We perform a multi-stage training, by using one single refinement at a time. Such 
technique is used based on the complexity of single stage training and on the 
gradient propagation problems of training deeper architectures. The process 
consists of initializing the contractive side with the VGG weights. 
After that the multi-stage training begins and each refinement is trained until we 
reach the final stage that uses the first pooling layer.

We use Stochastic Gradient Descent with momentum as the optimizer, a mini 
batch of size one and a fixed learning 
rate of $1e^{-6}$. Based on the mini batch size we set the momentum to $0.99$, allowing us to use previous gradients as much as possible. Since the labels in our problem 
are unbalanced because the majority of the points belong to the non-movable 
class, we incorporate \textit{class balancing} as explained by \citeauthor{eigen2015predicting}~\cite{eigen2015predicting}. 

The output of the network is a pixel wise score $a^k_c$ for each class $c$. The required \textit{objectness} score $\xi^k\in[0,1]$ for a point $k$ is the posterior class probability for the \textit{movable} class.    
\\
\begin{equation}
\xi^k = \frac{\exp(a^k_1)}{\exp(a^k_1) + \exp(a^k_0)} 
\end{equation}


\subsection{RigidFlow}
In our previous work~\cite{dewan2016rigid}, we proposed a method for estimating pointwise motion in LiDAR scans. The input to our method are two consecutive scans and the output is the complete 6D motion for every point in the scan. The two main advantages of this method is that it allows estimation of different arbitrary motions in the scene, which is of critical importance when there are multiple dynamic objects in the scene and secondly it works for both rigid and non-rigid bodies.

We represent the problem using a factor graph $G= (\Phi,\mathcal{T},\mathcal{E})$ with two node types: factor nodes $\phi \in \Phi$ and state variables nodes $\tau_k \in \mathcal{T}$. Here, $\mathcal{E}$ is the set of edges connecting $\Phi$ and state variable nodes $\mathcal{T}$. 

The factor graph describes the factorization of the function
\begin{equation}
\phi(\mathcal{T}) = \prod_{i\in I_d}\phi_d(\tau_i) \prod_{l\in N_p} \phi_p(\tau_i,\tau_j),
\label{eq:factor_graph}
\end{equation}

where $\mathcal{T}$ is the following rigid motion field:
\begin{equation}
\mathcal{T} = \lbrace \tau_k \mid \tau_k\in \mathit{SE}(3), k=1,\ldots,K \rbrace
\label{eq:motion_set}
\end{equation}

$\lbrace \phi_d, \phi_p\rbrace \in \phi$ are two types of factor nodes describing the energy potentials for the data term  and regularization term  respectively. The term $I_d$ is the set indices corresponding to keypoints in the first frame and $N_p=\lbrace \langle 1,2 \rangle, \langle 2,3 \rangle,\ldots, \langle i,j\rangle \rbrace$ is the set containing indices of neighboring vertices. The data term, defined only for keypoints is used for estimating motion, whereas the regularization term asserts that the problem is well posed and spreads the estimated motion to the neighboring points.
The output of our method is a dense rigid motion field $\mathcal{T}^\ast$, the solution of the following energy minimization problem:
\begin{equation}
\mathcal{T}^\ast = \operatorname*{arg\,min}_\mathcal{T} E(\mathcal{T}), 
\end{equation}
where the energy function is:
\begin{equation}
 E(\mathcal{T})=- \ln \phi(\mathcal{T}) 
\label{eq:energy_fucntion} 
\end{equation}
A more detailed explanation of the method is presented by \citeauthor{dewan2016rigid}~\cite{dewan2016rigid}.

\subsection{Bayes Filter for Semantic Classification}
\label{sec:Bayes filter}
The rigid flow approach estimates pointwise motion, however it does not provide the semantic level information. To this end, we propose a Bayes filter method for combining the learned semantic cues from the neural network with the motion cues for classifying a point as non-movable,  movable and dynamic. The input to our filter is
the estimated 6D motion, odometry and the \textit{objectness} score from the neural network. The \textit{dynamicity} score is calculated within the framework by comparing the motion with the odometry.

The \textit{objectness} score from the neural network is sufficient for classifying points as movable and non-movable, however, we still include this information in filter framework for the following two reasons:

\begin{itemize}
\item Adding object level information improves the results for dynamic classification because a point belonging to a non-movable object has infinitesimal chance of being dynamic, in comparison to a movable object. 
\item The current neural network architecture does not account for the sequential nature of the data. Therefore, having a filter over the classification from the network, allows filtering of wrong classification results by using the information from the previous frames. The same holds for classification of dynamic points as well.
\end{itemize}

\noindent
For every point $P^k_t\in \mathbb{R}^3$ in the scan, we define a state variable $x_t = \lbrace \texttt{dynamic, movable, non-movable} \rbrace$. The objective is to estimate the belief of the current state for a point $P^k_t$.

\begin{equation}
Bel(x^k_t) = p(x_t^k \mid x^k_{1:t-1}, \tau^k_{1:t},\xi^k_{1:t},o_t^k)\label{eq:bayes_filter}
\end{equation}
The current belief depends on the previous states $x_{1:t-1}^k$, motion measurements $\tau^k_{1:t}$, object measurements $\xi^k_{1:t}$ and a Bernoulli distributed random variable $o^k_t$. This variable models the object information, where $o^k_t = 1$ means that a point belongs to an object and therefore it is movable. For the next set of equations we skip the superscript $k$ that represents the index of a point.

\begin{align}
Bel&(x_t) = p(x_t \mid x_{1:t-1}, \tau_{1:t}, o_t,\xi_{1:t})\\
        & = \eta p(\tau_t,o_t\mid x_t,\xi_{1:t})\int p(x_t\mid x_{t-1})Bel(x_{t-1})dx_{t-1}\label{eq:bayes_filter_factor}\\
        & = \eta p(\tau_t\mid x_t)p(o_t\mid x_t,\xi_{1:t}) \int p(x_t\mid 
x_{t-1})Bel(x_{t-1})dx_{t-1}\label{eq:object_measure}
\end{align}     

In~\eqref{eq:bayes_filter_factor} we show the simplification of the~\eqref{eq:bayes_filter} using the Bayes rule and the Markov assumption. The likelihood for the motion measurement is defined in~\eqref{eq:motion_like}.
\begin{equation}
p(\tau_t\mid x_t) = \mathcal{N}(\tau_t;\hat{\tau_t},\Sigma)
\label{eq:motion_like}
\end{equation}
It compares the expected measurement $\hat{\tau_t}$ with the observed motion. In our case the expected motion is the odometry measurement. The output of the likelihood function is the required \textit{dynamicity} score.


In~\eqref{eq:object_measure} we assume the independence between the estimated motion and the object information. To calculate the object likelihood we first update the value of the random variable $o_t$ by combining the current \textit{objectness} score $\xi_t$ with the previous measurements in a log-odds formulation (\eqref{eq:log_like}).
\begin{align}
l(o_t\mid\xi_{1:t}) = l(\xi_t) + l(o_t\mid\xi_{1:t-1}) - l(o_0)
\label{eq:log_like}
\end{align}

The first term on the right side incorporates the current measurement, the second term is the recursive term which depends on the previous measurements and the last term is the initial prior. In our experiments, we set $o_0 = 0.2$ because we assume that the scene predominately contains non-moving objects.

\begin{align}
 p(o_t\mid x_t,\xi_{1:t}) =
  \begin{cases}
    p(\neg o_t\mid \xi_{1:t})& \quad \text{if }  x_t = \texttt{non-movable}  \\
    p(o_t \mid \xi_{1:t})& \quad \text{if }  x_t = \texttt{movable}  \\
    s\cdot p(o_t\mid \xi_{1:t})& \quad \text{if }  x_t = \texttt{dynamic}  \\
  \end{cases}
  \label{eq:object_like}
\end{align}
The object likelihood model is shown in~\eqref{eq:object_like}. 
As the neural network is trained to predict the non-movable and movable class, the first two cases in~\eqref{eq:object_like} are straightforward. For the case of dynamic object, we scale the prediction of movable class by a factor $s\in [0,1]$ since all the dynamic objects are movable, however, not all movable object are dynamic. This scaling factor approximates the ratio of number of dynamic objects in the scene to the number of movable objects. This ratio is environment dependent for instance on a \textit{highway}, value of $s$ will be close to $1$, since most of movable objects will be dynamic. For our experiments, through empirical evaluation, we chose the value of $s=0.6$.

\section{Results}
To evaluate our approach we use the dataset from the KITTI object benchmark and the dataset provided by \citeauthor{moosmann2013joint}~\cite{moosmann2013joint}. The first dataset provides object annotations but does not provide the labels for moving objects and for the second dataset we have the annotations for moving objects~\cite{dewan16icra}. Therefore to analyze the classification of movable and non-movable points we use the KITTI object benchmark and use the second dataset for examining the classification of dynamic points. For all the experiments, Precision and Recall are calculated by varying the confidence measure of the prediction. For object classification the confidence measure is the \textit{objectness} score and for dynamic classification the confidence measure is the output of the Bayes filter approach. The reported F1-score~\cite{naseer15ecmr} is always the maximum F1-score for the estimated Precision Recall curves and the reported precision and recall corresponds to the maximum F1-score.

\subsection{Object Classification}
Our classification method is trained to classify points on $cars$ as movable. The KITTI object benchmark provides $7481$ annotated scans. Out of these scans we chose $1985$ scans and created a dataset of $3789$ scans by tracking the labeled objects. The implementation of Fast-Net is based on a deep learning toolbox Caffe \cite{jia2014caffe}. The network was trained and tested on a system containing an NVIDIA Titan X GPU. For testing, we use the same validation set as mentioned by \citeauthor{chen2016multi}~\cite{chen2016multi}.

We provide quantitative analysis of our method for both pointwise prediction and object-wise prediction. For object-wise prediction we compare with these methods~\cite{XiaozhiCVPR16,chen20153d,li2016vehicle,chen2016multi}. Output for all of these methods is bounding boxes for the detected objects. A direct comparison with these methods is difficult since output of our method is pointwise prediction, however, we still make an attempt by creating bounding boxes out of our pointwise prediction as a post-processing step. We project the predictions from 2D image space to a 3D point cloud and then estimate 3D bounding boxes by clustering points belonging the to same surface as one object. The clustering process is described in our previous method~\cite{dewan2016rigid}.
\begin{table}[h!]
 \centering
 \caption{Object Classification $AP_{3D}$}
\begin{tabular}{|c|C{2cm}|C{1cm}|C{1.2cm}|C{1cm}|C{0.8cm}|}
\hline
\multirow{2}{*}{Method} & \multirow{2}{*}{Data }&\multicolumn{3}{ c |}{IoU=0.5}&\multirow{2}{*}{time}\\\cline{3-5}
 & & Easy & Moderate & Hard & \\
 \hline
 Mono3D~\cite{XiaozhiCVPR16}& Mono&25.19&18.2&15.52&4.2s\\
 3DOP~\cite{chen20153d}&Stereo&46.04&34.63&30.09&3s\\
 VeloFCN~\cite{li2016vehicle}&LiDAR&67.92&57.57&52.56&1s\\
 MV3D~\cite{chen2016multi}&LiDAR (FV)&74.02&62.18&57.61&-\\
 MV3D~\cite{chen2016multi}&LiDAR (FV+BV)&95.19&87.65&80.11&0.3s\\
 MV3D~\cite{chen2016multi}&LiDAR (FV+BV+Mono)&\textbf{96.02}&\textbf{89.05}&\textbf{88.38}&0.7s\\
 Ours &LiDAR&95.31 & 71.87 &70.01&\textbf{0.07s}\\
 \hline
\end{tabular}
  \label{tab:object_bbox}
\end{table}
\begin{table}[h!]
 \centering
 \caption{Pointwise versus Object-wise Prediction}
  \begin{tabular}{|c|C{1cm}|C{1.5cm}|C{2.5cm}|C{1.5cm}|}
  \hline
  Method&Recall&Recall (easy)&Recall (moderate)&Recall (hard)\\
  \hline
  pointwise&\textbf{81.29}&\textbf{84.79}&\textbf{71.07}&\textbf{68.10}\\
  object-wise&60.47&87.91&48.44&46.80\\
  \hline
  \end{tabular}
 \label{tab:pointvsobject}
\end{table}
\begin{table}[h!]
 \centering
 \caption{Object Classification F1-score}
 \begin{tabular}{ |c|C{1.5cm}|C{1.5cm}|C{1.5cm}|}
 \hline
 Method&F1 Score&Precision&Recall\\
 Ours(Fast-Net)&\textbf{80.16}&\textbf{79.06}&\textbf{81.29}\\
 Seg-Net&69.83&85.27&59.12\\
 Without \textit{Class Balancing} &78.14&76.73&79.60\\
 \hline
 \end{tabular}
  \label{tab:segnet}
\end{table}
 \begin{figure}[t]
  \centering
  \begin{tabular}{cc}
    \includegraphics[width=0.45\columnwidth]{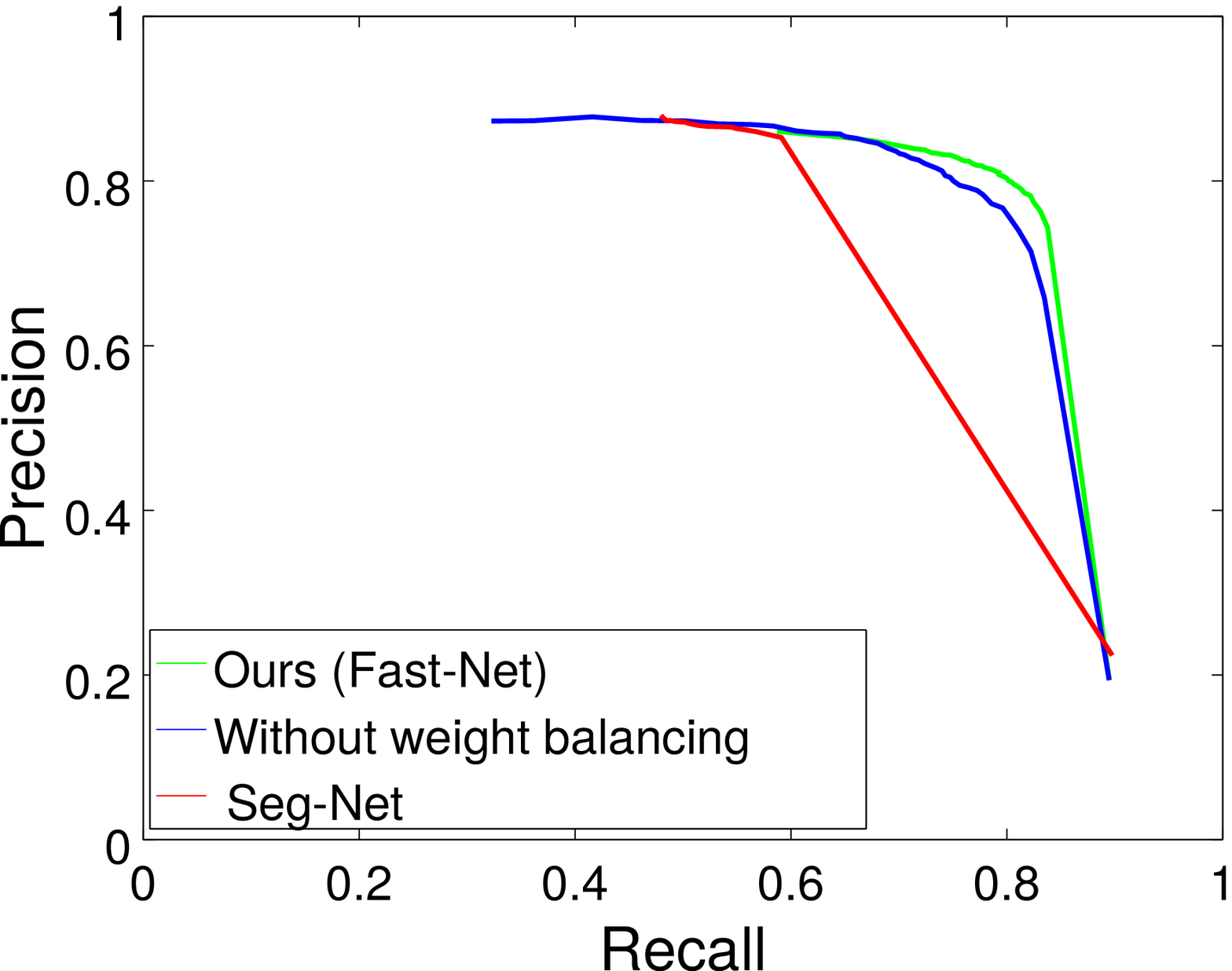}
    \includegraphics[width=0.45\columnwidth]{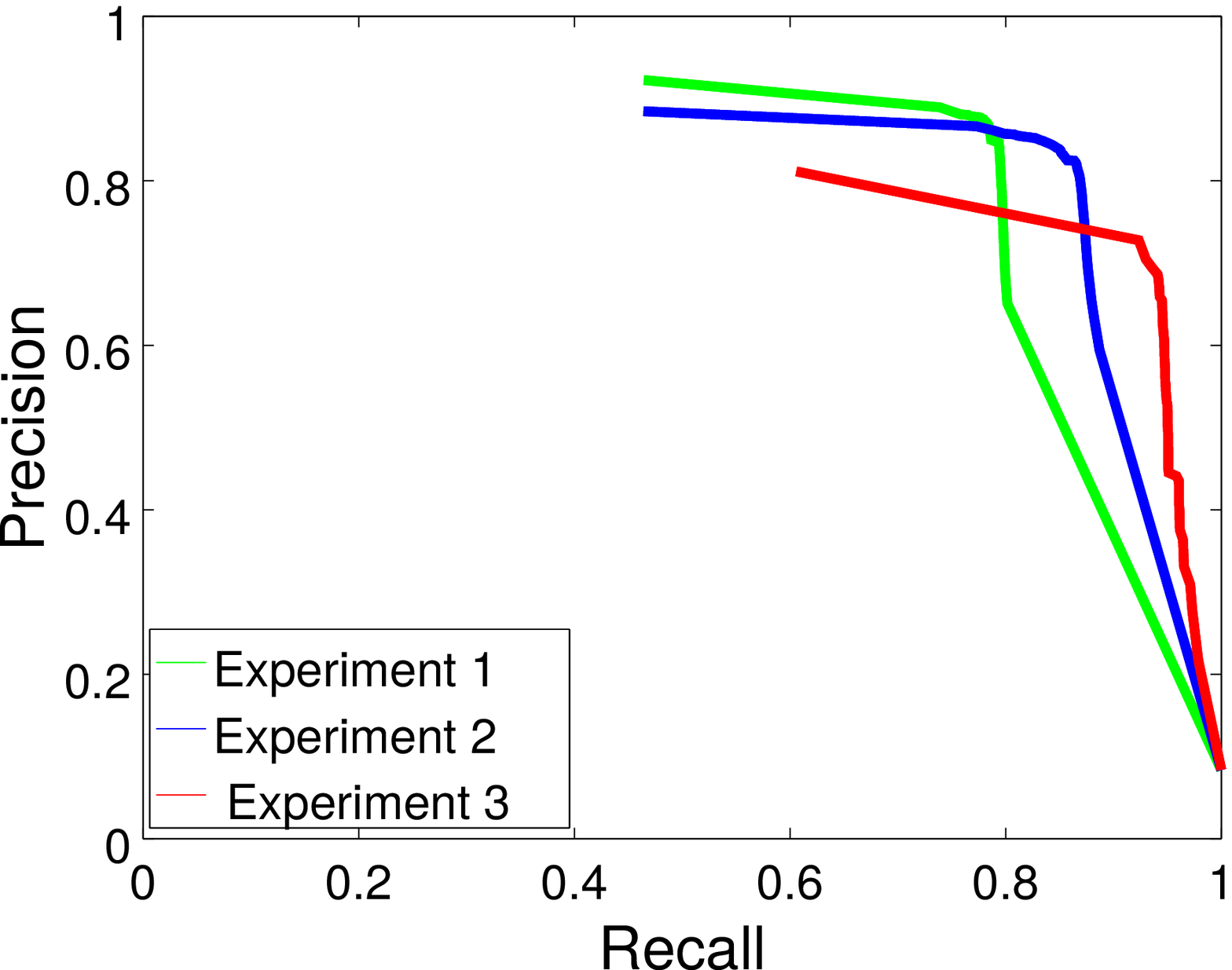}
  \end{tabular}
    \caption{Left: Precision Recall curves for object classification. Right: Precision recall curves for dynamic classification. Experiment 1 involves using the proposed Bayes filter approach, in experiment 2, the object information is not updated recursively and for the experiment 3, only motion cues are used for classification.}
    \label{fig:pr}
\end{figure}
\begin{figure*}[ht!]
  \centering
    \fbox{\includegraphics[width=0.85\textwidth]{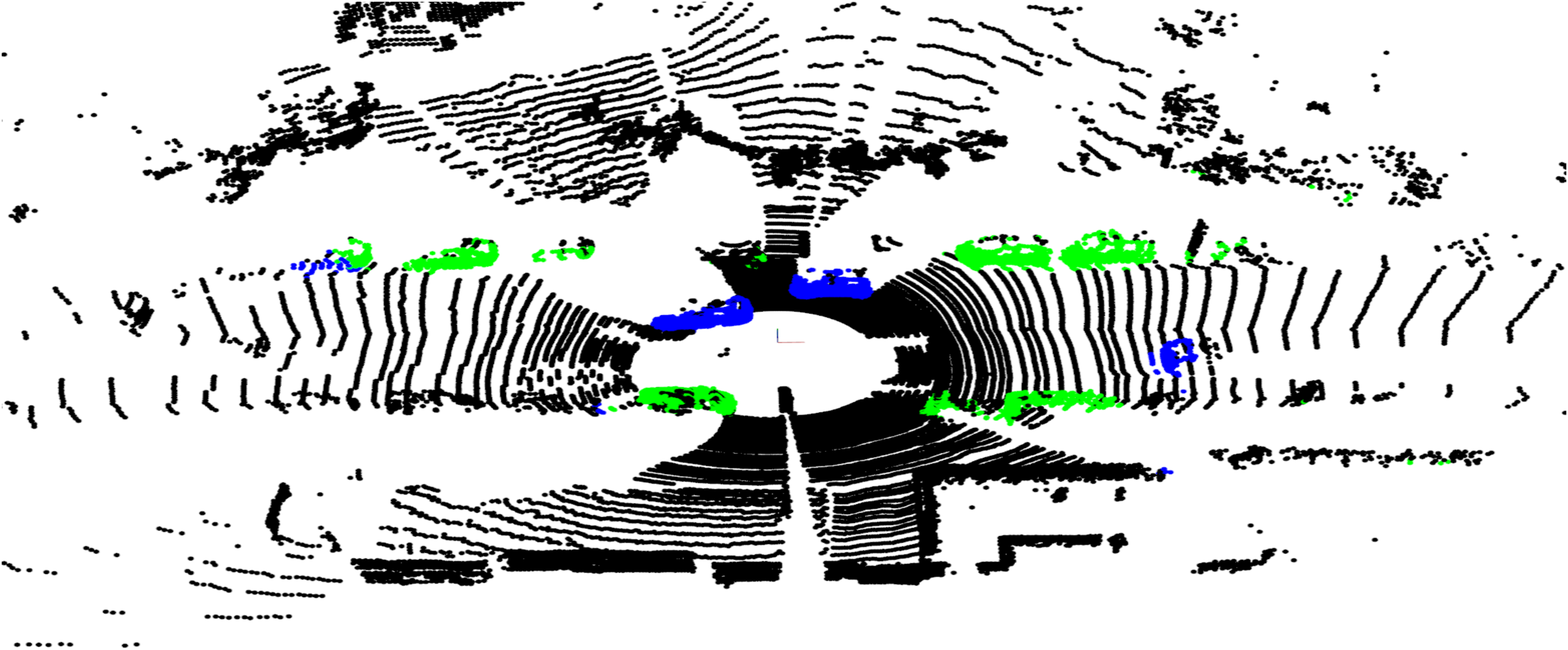}}
    \caption{A LiDAR scan from Scenario-B. Points classified as non-movable are shown in black, points classified as movable are shown in green and points classified as dynamic are shown in blue.}
    \label{fig:big_image}
 \end{figure*}

For object-wise precision, we follow the KITTI benchmark guidelines and report average precision for easy, moderate and hard cases. The level of difficulty depends on the height of the ground truth bounding box, occlusion level and the truncation level. We compare the average precision for 3D bounding boxes $AP_{3D}$ and the computational time with the other methods in~\tabref{tab:object_bbox}. The first two methods are based on RGB image, third method is solely LiDAR based, and the last method combines multiple view points of LiDAR data with RGB data. Our method outperforms the first three methods and an instance of the last method (front view) in terms of $AP_{3D}$. The computational time for our method includes the pointwise prediction on a GPU and object-wise prediction on CPU. The time reported for all the methods in~\tabref{tab:object_bbox} is the processing time on GPU. The CPU processing time for object-wise prediction of our method is $0.30s$. Even though performance of our method is not comparable with the two cases where LiDAR front view (FV) data is combined with bird eye view (BV) and RGB data, the computational time for our method is nearly $10\times$ faster.

In~\tabref{tab:pointvsobject} we report the pointwise and object-wise recall for the complete test data and for the three difficulty levels. The object level recall correspond to the $AP_{3D}$ results in~\tabref{tab:object_bbox}. The reported pointwise recall is the actual evaluation of our method. The decrease in recall from pointwise prediction to object-wise is predominantly for moderate and hard case because objects belonging to these difficulty levels are often far and occluded therefore discarded during object clustering. The removal of small clusters is necessary because minimal over segmentation in image space potentially results in multiple bounding boxes in 3D space as neighboring pixels in 2D projected image can have large difference in depth, this is especially true for pixels on the boundary of an object. The decrease in performance from pointwise to object-wise prediction should not be seem as a drawback of our 
approach since our main focus is to estimate precise and robust pointwise prediction required for the semantic classification.\\

We show the Precision Recall curves for pointwise object classification in~\figref{fig:pr} (right). Our method outperforms Seg-Net and we report an increase in F1-score by 12\% (see~\tabref{tab:segnet}). This network architecture
was used by~\citet{fan2016semantic} in their approach. To highlight the significance of \textit{class balancing}, we trained a neural network without \textit{class balancing}. Inclusion of this information increases the recall predominantly at high confidence values (see~\figref{fig:pr}).


\subsection{Semantic Classification}

For the evaluation of semantic classification we use a publicly available dataset~\cite{moosmann2013joint}. In our previous work~\cite{dewan16icra} we annotated the dataset for evaluating moving object detection. The dataset consists of two sequences: Scenario-A and Scenario-B, of $380$ and $500$ frames of 3D LiDAR scans respectively.

\begin{figure}[t]
  \centering
    \fbox{\includegraphics[width=0.45\columnwidth]{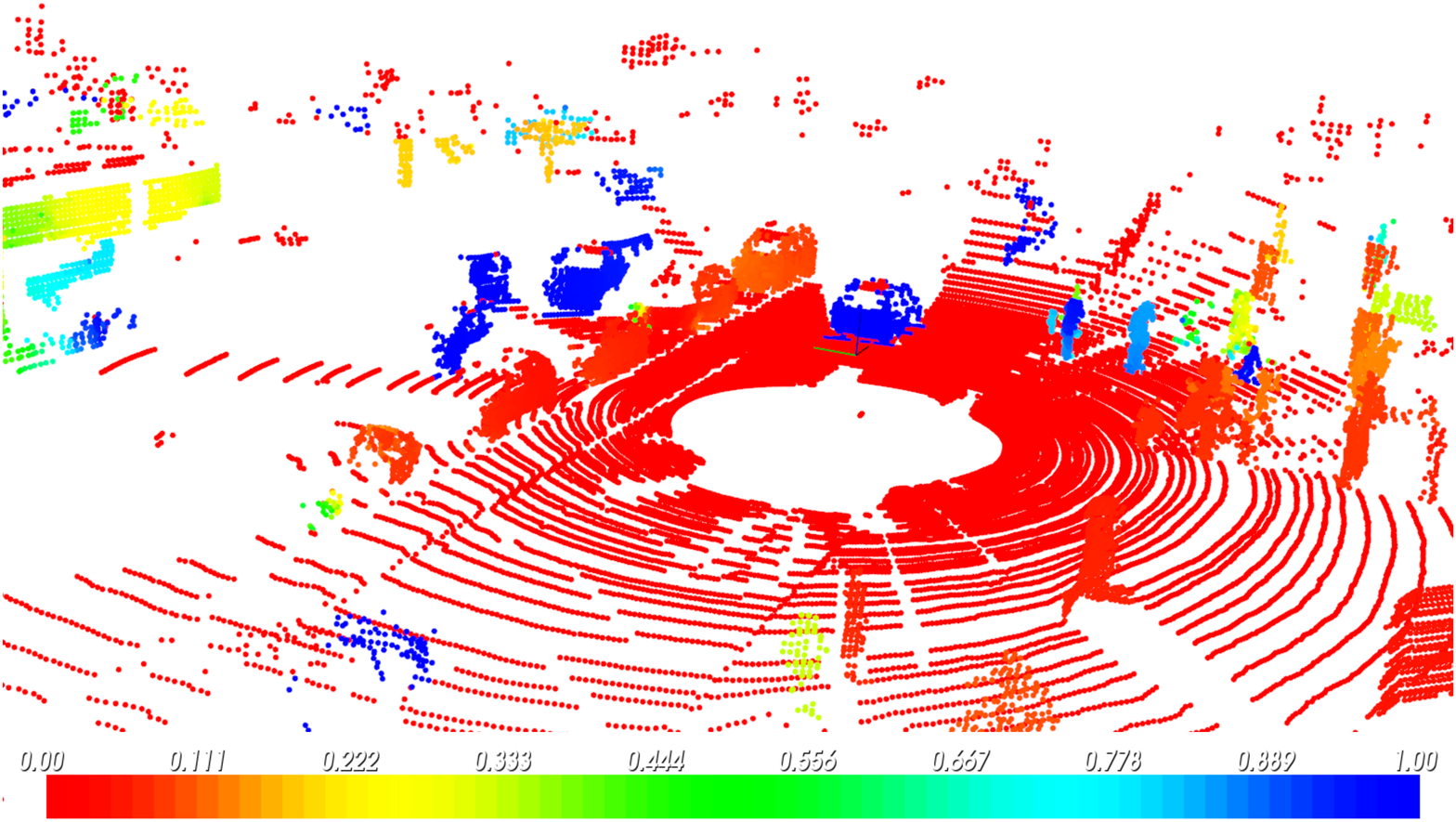}}
    \fbox{\includegraphics[width=0.45\columnwidth]{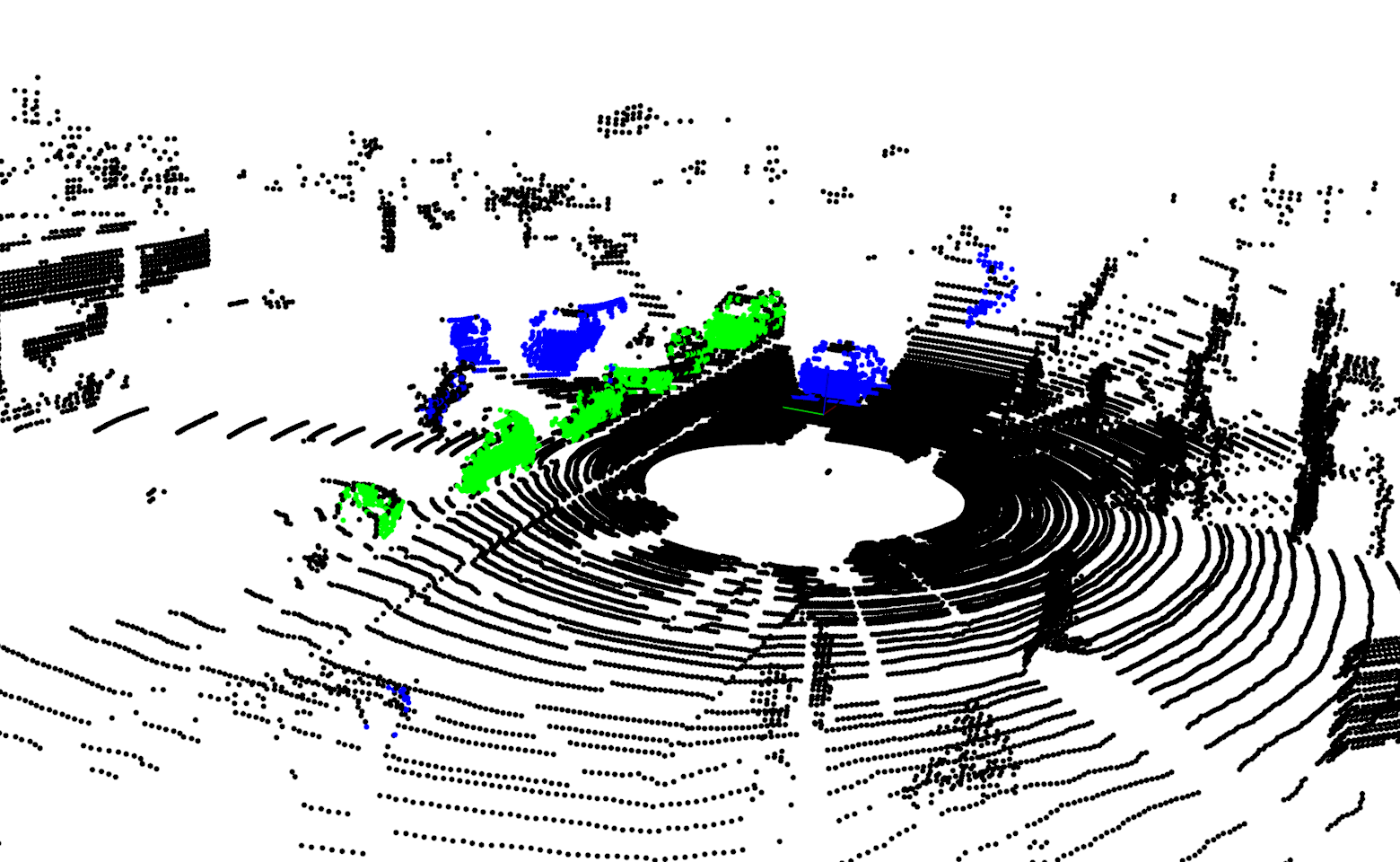}}
    \caption{Bayes filter correction for dynamic class. The image on the left is a visual representation of the \textit{dynamicity score} and right image is the classification results, where non-movable points are shown black, movable points are shown in green and dynamic points are shown in blue. The color bar shows the range of colors for the score ranging from $0$ to $1$. The misclassified points in the left image (left corner) are correctly classified as non-movable after the addition of the object information. 
    }
    \label{fig:bayes_filter}
\end{figure}

\begin{figure}[t]
  \centering
    \fbox{\includegraphics[width=0.28\columnwidth]{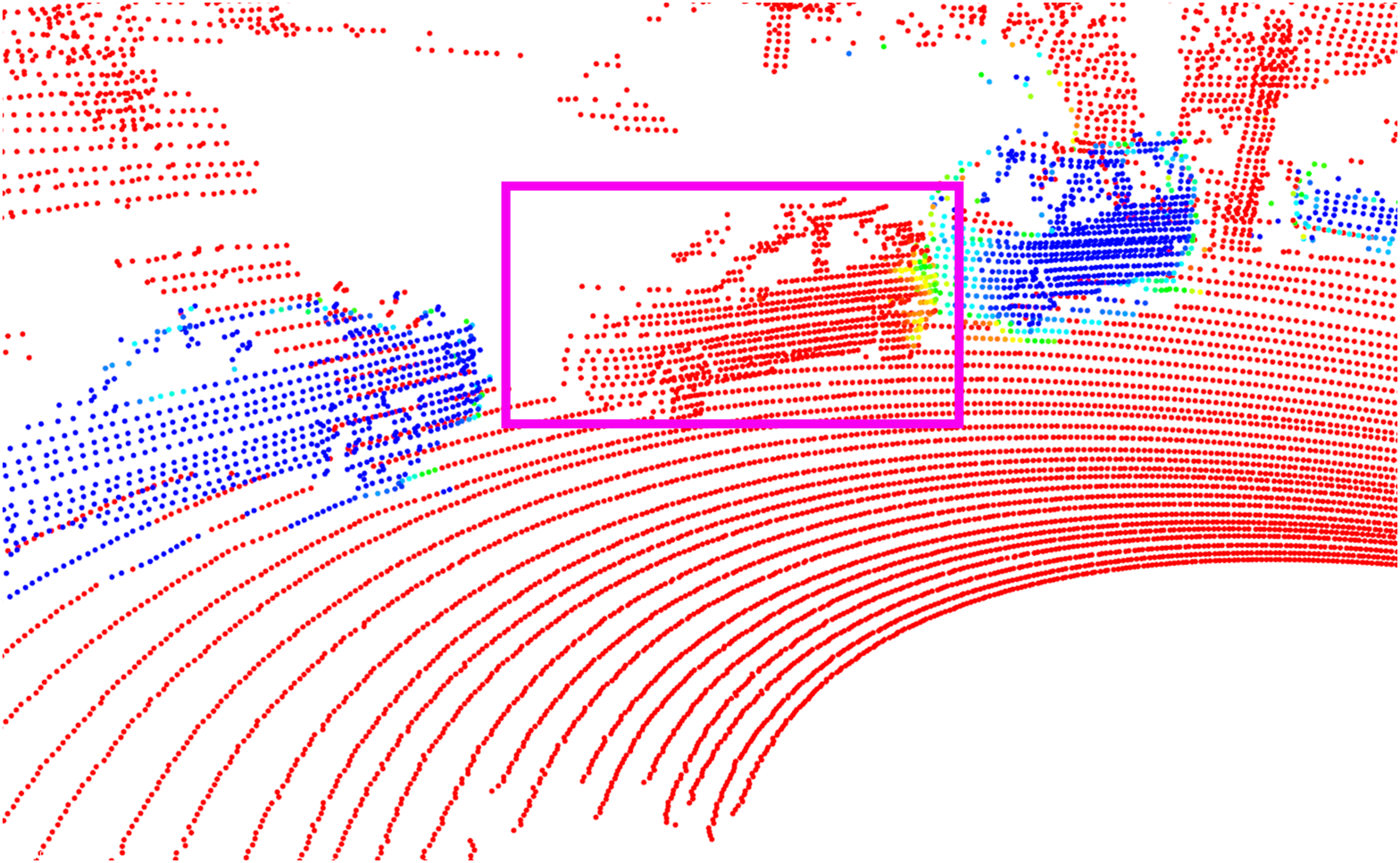}}
    \fbox{\includegraphics[width=0.28\columnwidth]{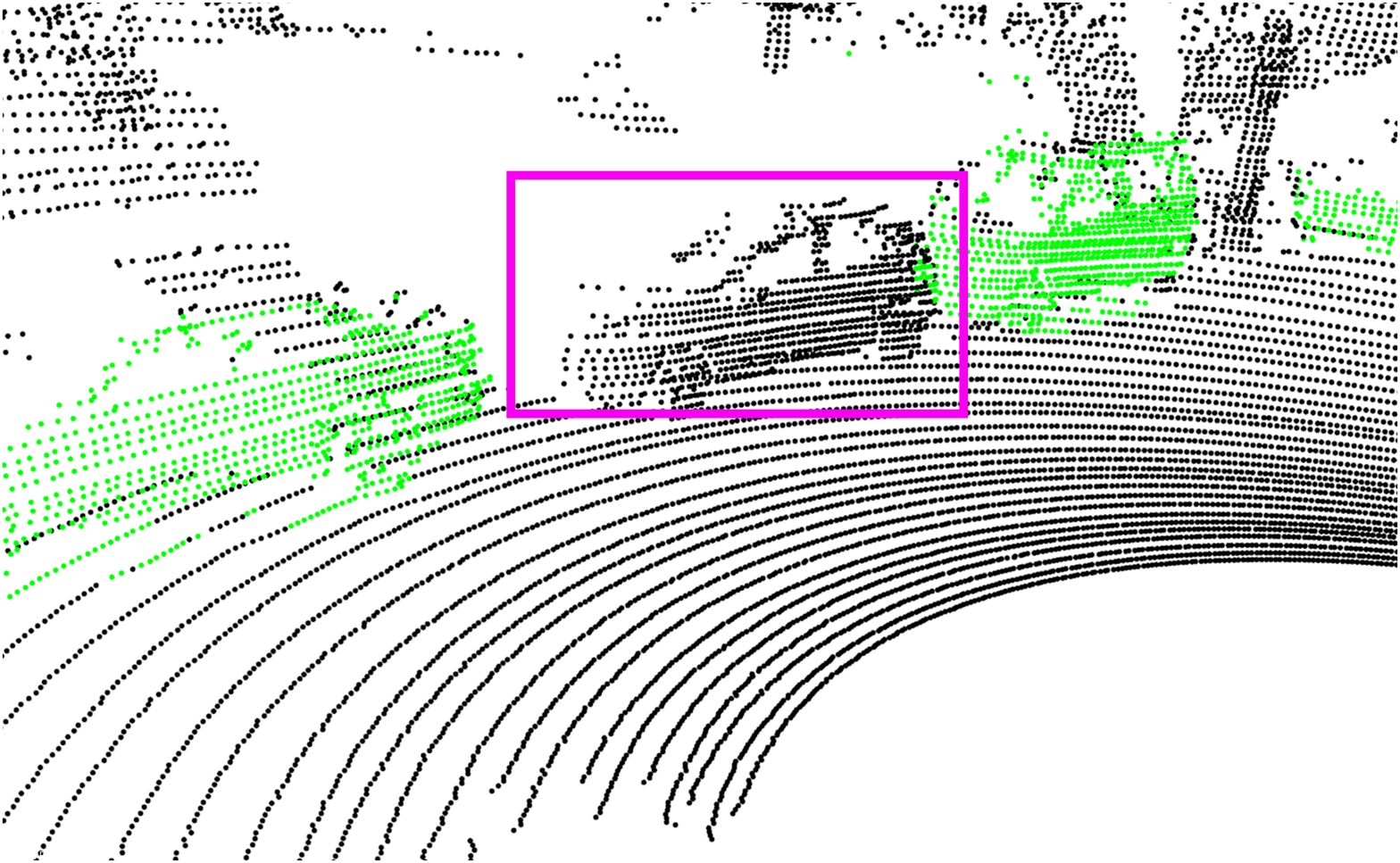}}
    \fbox{\includegraphics[width=0.28\columnwidth]{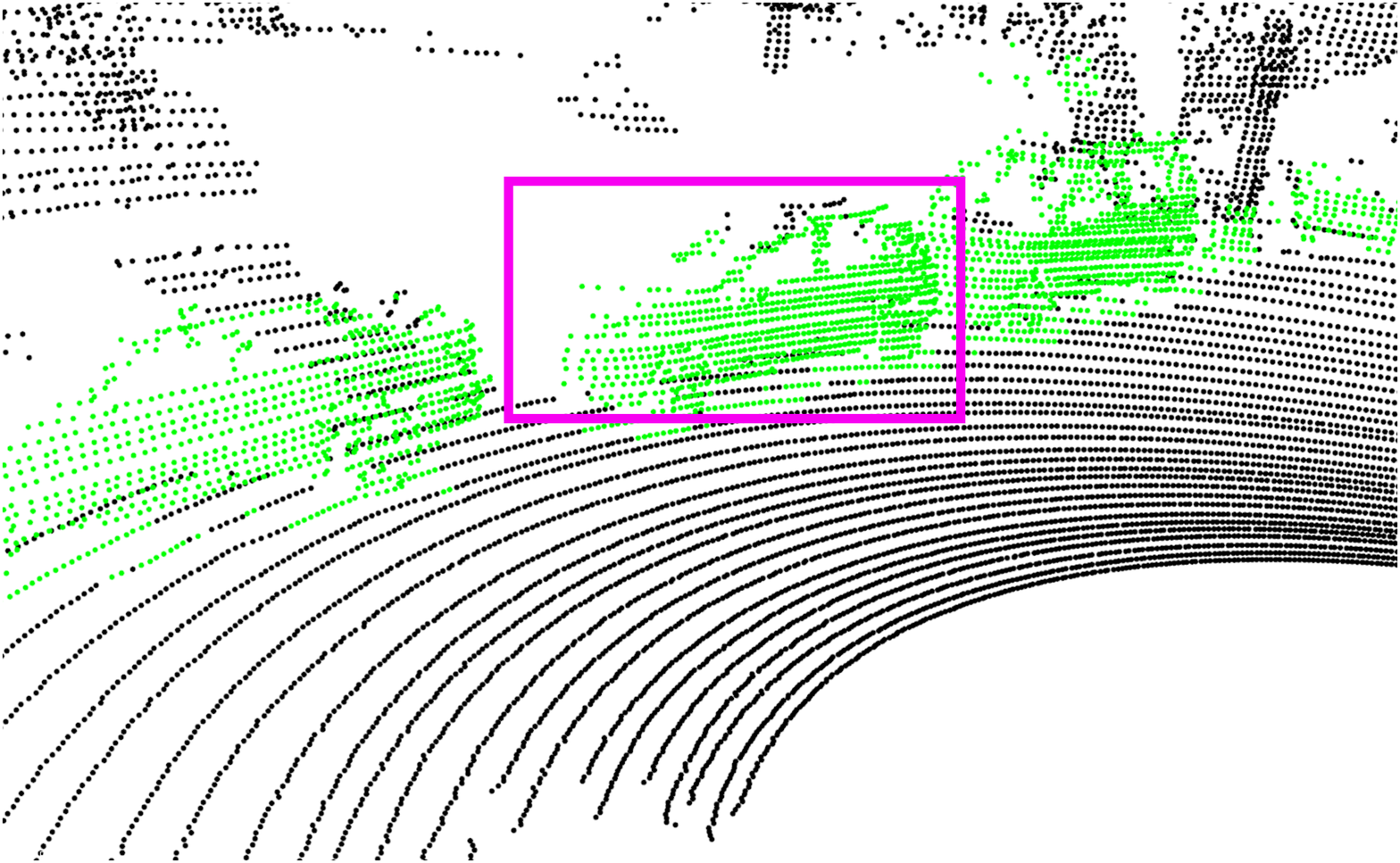}}
    
    \caption{Bayes filter correction for movable class. The image on the left is a visualization of \textit{objectness} score, where blue points corresponds to movable class. The purple rectangle highlights movable points with low score. The image in the middle and right shows classification of Bayes filter, where green points correspond to movable class. For the middle image, the object information was not recursively updated (second experiment) in the filter framework. For the right image we use the proposed Bayes filter approach (experiment 1). The points having low score is correctly classified as movable since the filter considers the current measurement and the previous measurements, which shows the
significance of updating the predictions from the neural network within the filter.}\label{fig:bayes_filter_object}
\end{figure}
\begin{table}[!h]
 \centering
 \caption{Dynamic Classification F1-score}
 \begin{tabular}{ |C{1.7cm}|C{1.0cm}|C{1.2cm}|C{1cm}|C{1.0cm}|C{1.2cm}|C{1cm}|}
 \hline
 \multirow{2}{*}{Method}&\multicolumn{3}{ c| }{Scenario A}&\multicolumn{3}{ c |}{Scenario B}\\\cline{2-7}
 &F1-score&Precision&Recall&F1-score&Precision&Recall\\
 Experiment 1&82.43&87.48&77.9&\textbf{72.24}&\textbf{76.29}&68.60\\
 Experiment 2&\textbf{84.44}&\textbf{83.85}&84.9&71.48&75.83&67.66\\
 Experiment 3&81.40&72.77&\textbf{92.34}&63.89&59.46&\textbf{69.62}\\ 
 \hline
 \end{tabular}
 \label{tab:bayes_filter}
\end{table}
We report the results for the dynamic classification for three different experiments. For first experiment we use the approach discussed in~\secref{sec:Bayes filter}. In the second experiment, we skip the step of updating the object information (see~\eqref{eq:log_like}) and only use the current \textit{objectness} score within the filter framework. For the final experiment, object information is not included in the filter framework and the classification of dynamic points rely solely on motion cues. 

We show the Precision Recall curves for classification of dynamic points for all the three experiments for Scenario-A in~\figref{fig:pr} (right). The PR curves illustrates that the object information affects the sensitivity (recall) of the dynamic classification, for instance when the classification is based only on motion cues (red curve), recall is better among all the three cases. With the increase in object information sensitivity decreases, thereby causing a decrease in recall. In~\tabref{tab:bayes_filter} we report the F1-score for all the three experiments on both the datasets. For both the scenarios, F1-score increases after adding the object information which shows the 
significance of leveraging the object cues in our framework. In~\figref{fig:bayes_filter}, we show a visual illustration for this case. 

For the Scenario-A, the highest score is for the second experiment. However, we would like to emphasize that the affect of including the predictions from the neural network in the filter is not only restricted to classification of dynamic points. In~\figref{fig:bayes_filter_object}, we show the impact of our proposed filter framework on the classification of movable points.

%
%
%
%

\section{Conclusion}
In this paper, we present an approach for pointwise semantic classification of a 3D LiDAR scan. Our approach uses an up-convolutional neural network for understanding the difference between movable and non-movable points and estimates pointwise motion for inferring the dynamics of the scene. In our proposed Bayes filter framework, we combine the information retrieved from the neural network with the motion cues to estimate the required pointwise semantic classification. We analyze our approach on a standard benchmark and report competitive results in terms for both, average precision and the computational time. Furthermore, through our Bayes filter framework we show the benefits of combining learned semantic information with the motion cues for robust and precise classification. For both the datasets
we achieve a better F1-score. We also show that introducing the object cues in the filter improves the classification of movable points.

%
%
%
%
%
%
%
%
\bibliographystyle{plainnat}
\footnotesize
\bibliography{dewan17iros}

\end{document}